\documentclass[runningheads]{llncs}
\usepackage{graphicx}
\usepackage{amsmath}
\usepackage{xcolor}
\usepackage{array}
\usepackage{multirow}
\usepackage{subfigure}
\usepackage{amsfonts}
\usepackage{ulem}
\usepackage{url}
\usepackage[misc]{ifsym}

\begin{document}
%
\title{Exploring Graph-aware Multi-View Fusion for Rumor Detection on Social Media}
\titlerunning{Graph-aware Multi-View Fusion for Rumor Detection}
%
\author{Yang Wu\inst{1,2} \and
\Letter{Jing Yang}\inst{1} \and
Xiaojun Zhou\inst{1} \and
Liming Wang\inst{1} \and
Zhen Xu\inst{1}}
\authorrunning{Y. Wu et al.}
%
\institute{ Institute of Information Engineering, Chinese Academy of Sciences\and
School of Cyber Security, University of Chinese Academy of Sciences\\
\email{\{wuyang0419, yangjing, zhouxiaojun, wangliming, xuzhen\}@iie.ac.cn}}

\toctitle{Exploring Graph-aware Multi-View Fusion for Rumor Detection on Social Media}
\tocauthor{Yang Wu, Jing Yang, Xiaojun Zhou, Liming Wang and Zhen Xu}
\index{Wu, Yang}
\index{Yang, Jing}
\index{Zhou, Xiaojun}
\index{Wang, Liming}
\index{Xu, Zhen}
  
\maketitle              
\begin{abstract}
Automatic detecting rumors on social media has become a challenging task. Previous studies focus on learning indicative clues from conversation threads for identifying rumorous information. 
However, these methods only model rumorous conversation threads from various views but 
fail to fuse multi-view features very well. 
In this paper, we propose a novel multi-view fusion framework for rumor representation learning and classification. It encodes the multiple views based on Graph Convolutional Networks (GCN), and leverages Convolutional Neural Networks (CNN) to capture the consistent and complementary information among all views and fuse them together. Experimental results on two public datasets demonstrate that our method outperforms state-of-the-art approaches.

\end{abstract}
\section{Introduction}
Social media has become an essential platform for people to obtain and share information. While bringing convenience, it also gives us too many dire challenges, one of which is the proliferation of rumors. 
A \textit{rumor} is generally defined as a statement that emerges and spreads among people whose truth value is true, unverified, or false \cite{difonzo2007rumor,qazvinian2011rumor,ma2018rumor,li2019rumor}.
Because social media lacks effective authentication techniques for user-generated content, users publish rumors without scruples, significantly reducing global Internet information credibility.
Social psychology research results show that humans are only slightly higher than 50\% (55\%-58\%) in their ability to identify deceptive information \cite{rubin2010deception}, which means that people are very easily deceived by false rumors.
Facing the massive amount of information on social media, it seems pretty powerless to verify rumors manually. Therefore, it is necessary to develop an automatic and assistant approaches to debunk rumors on social media.

\begin{figure}[htp]
	\centering  
	\subfigure[A false rumor]
	{
		\begin{minipage}[t]{0.22\linewidth}
			\centering
			\includegraphics[width=\textwidth]{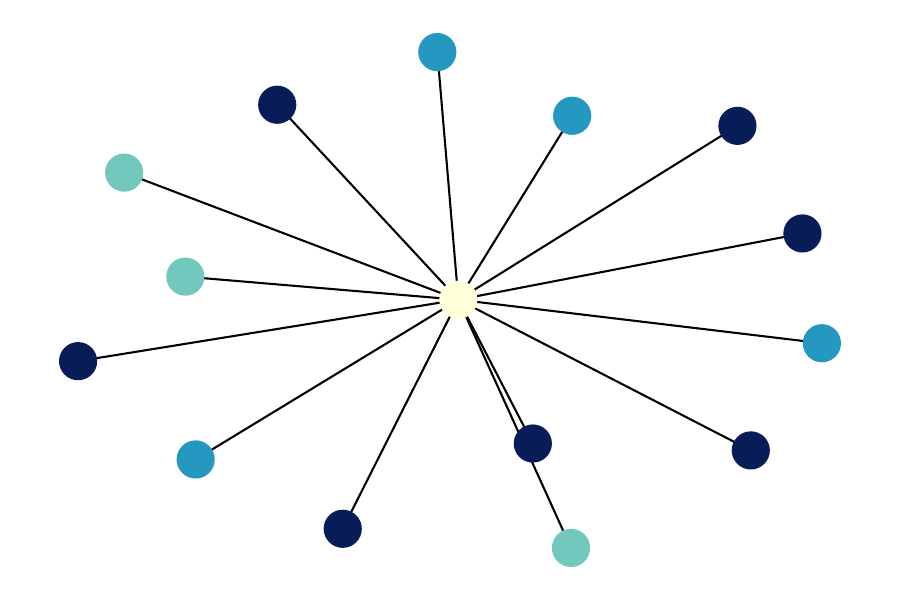}
		\end{minipage}
	}
	\subfigure[A false rumor]
	{
		\begin{minipage}[t]{0.22\linewidth}
			\centering
			\includegraphics[width=\textwidth]{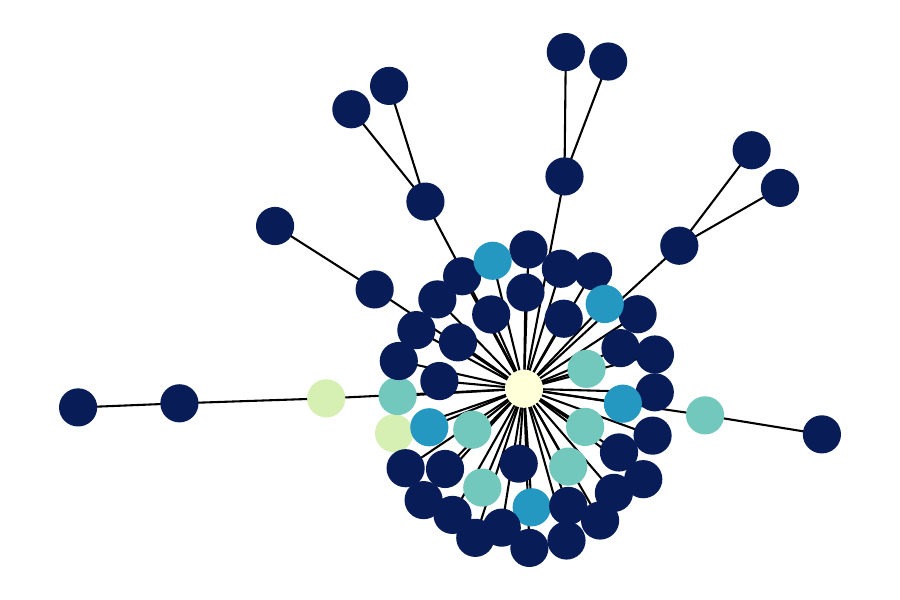}
		\end{minipage}
	} 
	\subfigure[A true rumor]
	{
		\begin{minipage}[t]{0.22\linewidth}
			\centering
			\includegraphics[width=\textwidth]{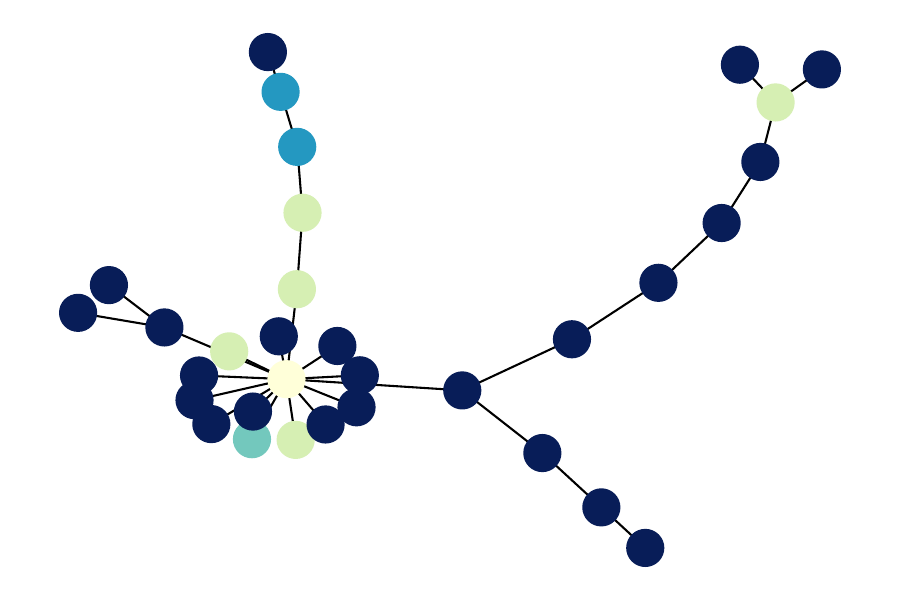}
		\end{minipage}
	}
	\subfigure[A true rumor]
	{
		\begin{minipage}[t]{0.22\linewidth}
			\centering
			\includegraphics[width=\textwidth]{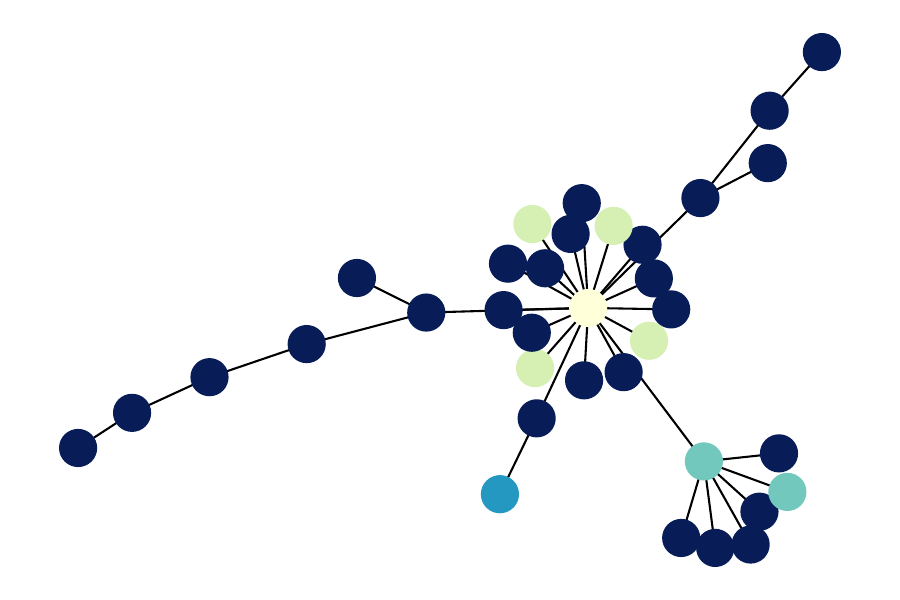}
		\end{minipage}
	}
	\caption{Conversation threads of four rumorous source posts from dataset, SemEval-2017. Nodes may express stances on their parents or source node, such as supporting ({\color[RGB]{216,241,180}$\bullet$}), denying ({\color[RGB]{149,219,201}$\bullet$}), querying ({\color[RGB]{53,177,208}$\bullet$}), and commenting ({\color[RGB]{7,64,78}$\bullet$}). The bright yellow nodes ({\color[RGB]{255,250,205}$\bullet$}) denote source posts. Stance labels come from the dataset SemEval-2017(refer to Section 5.1 for details)}
	\label{f1}
\end{figure}

\textit{Rumor detection} in this paper is the task of automatically determining the veracity value of the source claim in a conversation thread \cite{li2019rumor}, which is the same as \textit{rumor verification} defined in some other studies \cite{kochkina2018all,wei2019modeling,yu2020coupled}.
False (true) rumor means the veracity of the rumorous claim is false (true) \cite{ma2018rumor}.
As shown in Figure \ref{f1}, retweeted posts about the source claim form a conversation thread. 
On the topological structure, false rumors propagate shallowly around the root node, while true rumors present the characteristics of multi-point and multi-branch propagation, and the branches are long. 
A false rumor usually needs to attract more people's attention to spread it quickly, so its propagation is more likely to break out at the root node. In comparison, true rumors have no purpose of spreading, so their spreads are relatively scattered.
On the textual content, false rumors are more possibly to be denied or queried by most replies, while true rumors tend to be supported by people \cite{ma2018rumor}.
Therefore, the conversation threads under different source claims (false and true rumors) tend to exhibit different patterns in both topological and textual information. 

Besides textual information, existing studies mine the temporal, sequential, and structural properties of a rumorous conversation thread. Previous works model the conversation thread as a temporal model \cite{kochkina2018all,li2019rumor} or a sequential model \cite{lu2020gcan,khoo2020interpretable,yu2020coupled}, but ignores the structure information, which also reflects the behavior of rumors. Therefore, some studies model the rumor conversation as an undirected tree-structured graph and learn its structural features by GCN \cite{wei2019modeling} or Graph Attention Networks (GAT) \cite{lin2021rumor}. 
However, there are various relationships among nodes in a conversation thread, such as the top-down propagation relationship along retweet chains \cite{bian2020rumor,han2014energy} and bottom-up dispersion relationship within communities \cite{bian2020rumor,thomas2007lies} as shown in Figure \ref{f2}, which undirected GCN neglects. 
Ma et al. designed tree-structured Recursive Neural Networks (RvNN) \cite{ma2018rumor} to learn rumor features from top-down view. Nevertheless, their model only learns one pattern of rumors, lacking comprehensiveness.
Each type of relationship among nodes is considered as a view.
The top-down view indicates rumor propagation from the source post to forwarded posts along different paths. But it cannot represent information aggregation from leaf nodes to the root node across a social community, while the cases in the bottom-up view are on the contrary. So it is desirable to utilize the complementary information from both views.
Thus, recent studies use two GCNs to encode the top-down view and bottom-up view synchronously and obtain representations of two views by applying mean pooling over all nodes \cite{bian2020rumor,wei2021towards}.
However, they just concatenate two-view feature representations without considering node-to-node complementarity over both views. As shown in Figure \ref{f2} left, two views share the same set of nodes (i.e., posts). Although with the initial embedding value, the node embedding will be updated according to view-specific relationships. 
Intuitively, the way that learning representations of individual views followed by simple concatenation loses the global perspective. 
Just like identifying the shape of an object according to its three-view drawing, it is easy to make mistakes if observing three views separately.
Therefore, we think learning view-wise node representations and fusing them is the right way to make full use of consistent and complementary information in multiple views for rumor detection, which is still a challenge.

To achieve the aforementioned goal, we propose a novel \uline{G}raph-aware \uline{M}ulti-\uline{V}iew \uline{C}onvolutional Neural \uline{N}etwork (GMVCN) for rumor detection.
We are inspired by computer vision. A color image can be represented by red, green, and blue (RGB) color channels, and each pixel position has three values corresponding to RGB, respectively. The RGB channels are the three-color views of the image. 
The actual color of the image can only be seen by superimposing the three color views on each other.
In computer vision, the information of RGB views is effectively integrated by CNN for tasks such as image classification \cite{li2014medical}.
Inspired by this, we regard the rumor conversation thread as a color image and each node as a pixel. The top-down propagation view and bottom-up dispersion view are regarded as two channels of the image. 
Both views share the same set of nodes with initial values. 
In order to capture view-specific features, we leverage GCN to encode individual views, which updates the node embedding according to view-specific structures. Then a CNN-based sub-module is utilized to capture the consistent and complementary information between the two views and fuse them into a global rumor representation vector for prediction. 

The contributions of this work are as follows.
(1) We propose a novel graph-aware multi-view convolutional neural network to effectively integrate the multi-view information of a conversation thread for rumor detection. To the best of our knowledge, this is the first study to explore the multi-view fusion of conversation threads for rumor detection on social media.
(2) We innovatively treat a rumor conversation thread as an image, nodes as pixels, and multiple views as image channels. We utilize GCN to learn view-specific features and then explore to capture rumor characteristics from the consistent and complementary information of both views with CNN. 
(3) Experimental results on two public datasets show that our model outperforms several state-of-the-art approaches. Besides, we visualize what our proposed model has captured, and the results effectively verify that our method can learn inherent properties possessed by rumors on social media.


\section{Related Work}
\subsection{Rumor detection}
Most previous methods learns rumor characteristics from text contents, user profiles and retweet propagation \cite{castillo2011information,feng2012syntactic,chen2015misleading,yang2012automatic,kwon2013prominent,sampson2016leveraging}. Ma et al. \cite{ma2015detect} developed a time series model to capture the variation of social context information over time. Some studies model propagation patterns as tree structures based on kernel methods \cite{wu2015false,ma2017detect}, etc. However, these traditional approaches heavily rely on feature engineering, which is labor-intensive and limited. 

Recently, researchers explored applying deep neural networks to detect rumors. Ma et al. \cite{ma2016detecting} leveraged Recurrent Neural Networks (RNN) to learn the representations of relevant posts over spread time.
Liu et al. \cite{liu2018early} utilized both RNN and CNN to extract user profiles along the propagation time to determine the truth of posts. Lu et al. \cite{lu2020gcan} proposed a hybrid-based model which considers source tweets and user profiles. Yu et al. \cite{yu2020coupled} divided each long thread into shorter subthreads and learned local and global interactions among them based on a hierarchical Transformer framework. However, these approaches ignore the structure features of conversation threads, as shown in Figure \ref{f1}, which can reflect how posts are propagated over social media.

In order to capture indicative patterns from both text contents and propagation structures based on deep learning methods, Ma et al. \cite{ma2018rumor} proposed an RvNN-based model to learn hidden representations of tweets along propagation trees. 
Lin et al. \cite{lin2021rumor} enhanced the representation learning of each post and the interactions among them based on undirected graph neural Networks with multiple attention mechanisms. The tree-structured graphs encoded in these methods have only one edge type, but in reality, the relationships between nodes are various. Hence, Bian et al. \cite{bian2020rumor} focused on both top-down propagation relationship and bottom-up dispersion relationship among nodes for rumor detection. Wei et al. \cite{wei2021towards} removed unreliable relationships between nodes in rumor conversation threads based on Bian's work. However, the shortcoming of these works is that they cannot effectively integrate multiple views in rumor conversations to mine the difference between false rumors and true rumors from a global perspective.

\subsection{Multi-view Graph Learning}
In recent years, Graph Neural Networks have been adopted for various tasks according to their remarkable performance of representation learning on structured data, e.g., text classification \cite{zhang2020every}, recommendation system \cite{wei2021hierarchical}, etc. 
Representative graph neural networks include GCN \cite{kipf2016semi}, GAT \cite{velickovic2017graph}, and so on. However, these methods are tailored for single-view network representation \cite{xie2020mgat}. There are amount of multi-view networks in the real world, where each view consists of a type of relationship. Thus, substantial efforts have been dedicated to exploring multi-view graph learning \cite{zhang2018scalable,xu2019multi,xie2020mgat}. 
They mainly investigated how to integrate each view's node representations into a global node representation. Unlike them, we explore solving the problem of how to fuse features of multi-view graphs into a global graph feature representation vector for rumor detection. 



\section{Preliminaries}

\subsection{Problem Statement}
We define a rumor detection dataset as a set of conversation threads $\mathcal{C}=\left\{c_1,c_2,...,c_m\right\}$, where $c_i$ is the $i$-th conversation thread and $m$ is the number of conversation threads. Each thread $c_i$ consists of a source claim $r_i$ and a number of reply posts sorted in chronological order, which is denoted as $c_i=\left\{r_i,x_1^i, x_2^i,...,x_{n_i-1}^i,G_i\right\}$, where $x_j^i$ is the $j$-th relevant retweet, and $G_i$ is the propagation structure. Specifically, $G_i=\left\langle V_i,E_i\right\rangle$ is a tree-structured graph with the root node $r_i$, where $V_i=\left\{r_i,x_1^i,...,x_{n_i-1}^i\right\}$, and $E_i=\left\{e_{st}^i|s,t=0,...,n_i-1\right\}$ refers a set of edges that represents reply relationship among nodes. For example, if $x_2^i$ has a response to $x_1^i$, there will be an directed edge $x_1^i \to x_2^i$ in $E_i$, i.e., $e_{12}^i$.

The rumor detection task in our paper is formulated as a supervised classification problem, aiming to learn a classification function $f:\mathcal{C} \to \mathcal{Y}$, where $\mathcal{C}$ and $\mathcal{Y}$ are the sets of conversations and labels, respectively.

\subsection{Graph Convolutional Networks}
In recent years, graph convolutional networks have been demonstrated superior performance for dealing with graph data in a variety of NLP tasks, such as text classification \cite{yao2019graph}, modeling knowledge graph \cite{tian2020ra}, and recommendation system \cite{he2020lightgcn}, etc. GCNs capture information from a node's direct and indirect neighbors by multiple Graph Convolution Layers (GCL) and then update the node's representation. Given the adjacency matrix ${\rm \mathbf{A}} \in \mathbb{R}^{N\times N}$ for a graph $G$. Its layer-wise propagation rule is:
\begin{equation}\label{eq1}
    {\rm \mathbf{H}}_{l+1}=\sigma(\hat{\rm \mathbf{A}}{\rm \mathbf{H}}_l{\rm \mathbf{W}}_l)
\end{equation}
where $l$ is the layer number, $\hat{\rm \mathbf{A}}=\tilde{\rm \mathbf{D}}^{-\frac{1}{2}}\tilde{\rm \mathbf{A}}\tilde{\rm \mathbf{D}}^{-\frac{1}{2}}$ the normalized symmetric weight matrix ($\tilde{\rm \mathbf{A}}={\rm \mathbf{A}}+{\rm \mathbf{I}}_N$, $\tilde{\rm \mathbf{D}}_{ii}=\sum_{j}\tilde{\rm \mathbf{A}}_{ij}$, where ${\rm \mathbf{I}}_N$ is the identity matrix), ${\rm \mathbf{W}}_l \in \mathbb{R}^{D\times D}$ is a layer-specific trainable weight matrix, $\sigma(\cdot)$ is an activation function, e.g., the ReLU function, ${\rm \mathbf{H}}_l\in \mathbb{R}^{N \times D}$ is the hidden feature matrix in the $l^{th}$ layer.

\section{Proposed Method}

\begin{figure}[htp]
  \centering
  \includegraphics[width=\linewidth]{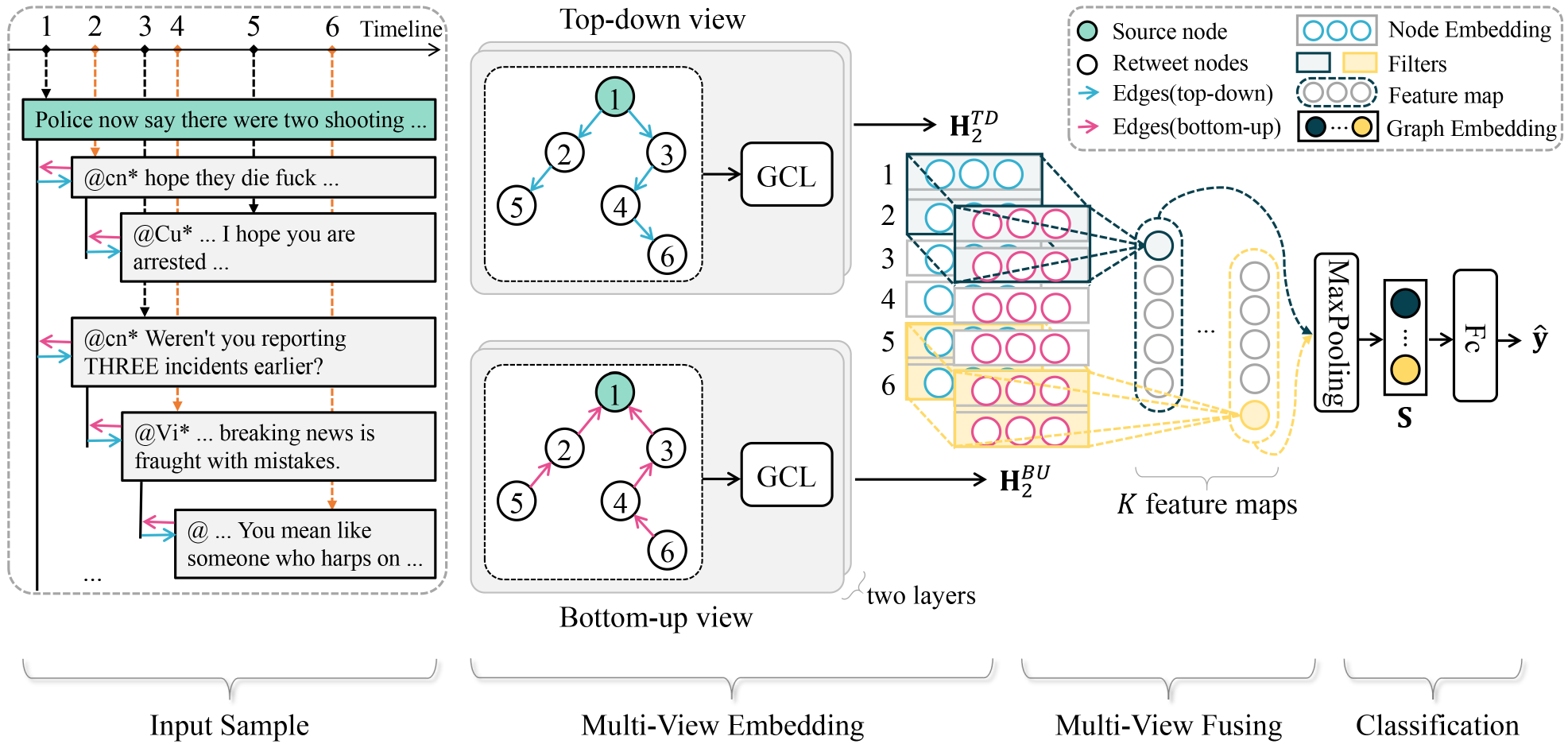}
  \caption{The architecture of our graph-aware multi-view convolutional neural networks}
  \label{f2}
\end{figure}

In this section, we introduce our proposed Graph-aware Multi-View Convolutional neural Network model that learns suitable high-level representations from rumor conversation threads. Figure \ref{f2} illustrates an overview of our model, which consists of three main components, including multi-view embedding, multi-view fusing, and classification. We will depict how to apply the GMVCN model to determine the veracity value of the source post $r_i$ in the conversation thread $c_i$. To better present our method, we omit the subscript $i$ in the following content. The other conversations are calculated in the same manner.

\subsection{Multi-View Embedding}
Given $G=\left\langle V,E\right\rangle$, we transform the edge set $E$ to an adjacency matrix ${\rm \mathbf{A}}\in \mathbb{R}^{n\times n}$, where ${\rm \mathbf{A}}_{ij}=1$ if node $x_j$ directly replies the node $x_i$. Our model focuses on two different views of the conversation thread, the top-down propagation view and the bottom-up dispersion view, which have opposite direct edges. ${\rm \mathbf{A}}$ only contains edges in the top-down view from the parent node to its children nodes. In order to reduce overfitting of our model, we apply the Dropedge technique \cite{DBLP:journals/corr/abs-1907-10903} on $G$ as previous research \cite{bian2020rumor,wei2021towards}, which randomly drops out $p$ percentage of edges. And then we obtain a new adjacency matrix ${\rm \mathbf{A}}^{\prime}$, so the adjacency matrix of the top-down view is represented as ${\rm \mathbf{A}}^{TD}={\rm \mathbf{A}}^{\prime}$. For bottom-up view, the adjacency matrix is represented as ${\rm \mathbf{A}}^{BU}={{\rm \mathbf{A}}^{\prime}}^{\rm T}$. The nodes of two views share the same initial feature matrix ${\rm \mathbf{X}}$, where the top-5000 words represent features of each node in terms of the TF-IDF values. Then their embeddings are updated by two different GCNs, respectively. The graph convolution operation in one GCN layer is depicted as Eq. (\ref{eq1}). For top-down view, we feed ${\rm \mathbf{A}}^{TD}$ and ${\rm \mathbf{X}}$ to two-layer GCN as follows:
\begin{align}
    {\rm \mathbf{H}}_1^{TD}=\sigma(\hat{\rm \mathbf{A}}^{TD}{\rm \mathbf{X}}{\rm \mathbf{W}}_1^{TD}) \label{eq2} \\
    {\rm \mathbf{H}}_2^{TD}=\sigma(\hat{\rm \mathbf{A}}^{TD}{\rm \mathbf{H}}_1^{TD}{\rm \mathbf{W}}_2^{TD}) \label{eq3}
\end{align}
where ${\rm \mathbf{H}}_1^{TD}\in \mathbb{R}^{n \times d_1}$ and ${\rm \mathbf{H}}_2^{TD}\in \mathbb{R}^{n \times d_2}$ denote the hidden feature representations of the two-layer GCN. ${\rm \mathbf{W}}_1^{TD}\in \mathbb{R}^{d \times d_1}$ and ${\rm \mathbf{W}}_2^{TD}\in \mathbb{R}^{d_1 \times d_2}$ are convolution filter matrices shared by all nodes in the conversation thread. Here $\sigma(\cdot)$ is the ReLU function. To avoid overfitting, we apply Dropout to GCN layers. For bottom-up view, the calculation of two-layer GCN is similar to Eq. (\ref{eq2}) and Eq. (\ref{eq3}), which obtains ${\rm \mathbf{H}}_1^{BU}$ and ${\rm \mathbf{H}}_2^{BU}$, respectively.

\subsection{Multi-View Fusing}
After top-to-down and bottom-to-up graph embedding, each node $n_i$ has two representations ${\rm \mathbf{h}}_i^{TD} \in \mathbb{R}^{d_2}$ and ${\rm \mathbf{h}}_i^{BU}\in \mathbb{R}^{d_2}$ corresponding to two views. We treat each conversation thread as a color image, each view as a channel, and each node as a pixel. Therefore, the feature representations of the rumor conversation thread are formulated as follows:
\begin{equation}\label{eq4}
    {\rm \mathbf{H}}=
    \begin{bmatrix}
    {\rm \mathbf{h}}_1^{TD}, &  {\rm \mathbf{h}}_2^{TD}, & ..., & {\rm \mathbf{h}}_n^{TD} \\
    {\rm \mathbf{h}}_1^{BU}, &  {\rm \mathbf{h}}_2^{BU}, & ..., & {\rm \mathbf{h}}_n^{BU} \\
    \end{bmatrix}
    \in \mathbb{R}^{2\times n \times d_2}
\end{equation}

Given the multi-channel input ${\rm \mathbf{H}}$, a convolution operation, which involves multiple filters $\mathbf{W}_k \in \mathbb{R}^{2 \times l \times d_2}, k \in \begin{bmatrix}
1,K
\end{bmatrix}$, is applied to a window of $t$ nodes to produce new features according to the following operation:
\begin{equation}\label{eq5}
    {\rm \mathbf{s}}_i^k=\sigma(\mathbf{W}_k * {\rm \mathbf{H}}_{i:i+t-1}+{\rm \mathbf{b}})
\end{equation}
where $\mathbf{W}_k$ and ${\rm \mathbf{b}}$ are learnable parameter matrices, $*$ denotes multi-channel convolutional operations, and $\sigma(\cdot)$ is the ReLU activation function. We apply the filters to each possible window of $t$ nodes in ${\rm \mathbf{H}}$, and obtain the following feature map:
\begin{equation}\label{eq6}
    {\rm \mathbf{s}}^k=\begin{bmatrix}
        {\rm \mathbf{s}}_1^k, & {\rm \mathbf{s}}_2^k, & ..., & {\rm \mathbf{s}}_{n-t+1}^k
        \end{bmatrix}
\end{equation}

We then use max-over-time pooling operation on the above feature map and get the most important feature $\hat{\rm \mathbf{s}}^k=max\left\{{\rm \mathbf{s}}^k\right\}$ corresponding to this particular filter. All maximum values of the feature maps produced by all filters are concatenated as the final representation of the conversation thread:
\begin{equation}\label{eq7}
    {\rm \mathbf{S}}=\begin{bmatrix}
        \hat{\rm \mathbf{s}}^1, & \hat{\rm \mathbf{s}}^2, & ..., \hat{\rm \mathbf{s}}^K
    \end{bmatrix}
\end{equation}
where $K$ is the number of filters.

\subsection{Classification}
Finally, the learned representation of the conversation thread is fed to a full connection layer with softmax normalization for predicting the probability of the source post being a false rumor:
\begin{equation}\label{eq8}
    \hat{\rm \mathbf{y}}=softmax(FC({\rm \mathbf{S}}))
\end{equation}
The loss function is devised to minimize the cross-entropy value of the predictions and ground
truth distributions:
\begin{equation}\label{eq9}
    \mathcal{L}_\mathcal{C}=-\sum_{i}^{\vert \mathcal{Y} \vert}{\rm \mathbf{y}^i}log\hat{\rm \mathbf{y}}^i
\end{equation}
where ${\rm \mathbf{y}^i}$ denotes the vector representing distribution of
ground truth label for source claim in $c_i$.

\section{Experiments and Results}
In this section, we first evaluate the performance of GMVCN compared with several baselines on rumor detection. Then we further analyze our proposed model, i.e., the effect of every model component.

\subsection{Datasets}
To evaluate the effectiveness of our GMVCN model, we conduct experiments on two public real-world datasets, SemEval-2017 task 8\footnote{\url{https://alt.qcri.org/semeval2017/task8/index.php?id=data-and-tools}} and PHEME\footnote{\url{https://figshare.com/articles/dataset/PHEME_dataset_for_Rumour_Detection_and_Veracity_Classification/6392078}}. The statistics of two datasets are shown in Table \ref{t1}.

SemEval-2017 task 8 \cite{derczynski2017semeval} is composed of 325 rumorous conversation threads, which have been split into training, development, and test sets. These conversation threads are related to 10 events, and the test set covers two events data that the training and development set do not have. 
In addition to the rumor category label of each conversation thread, each post in the thread is annotated with its stance label (support, deny, query, or comment). Here stance labels only assist in analyzing the performance of our proposed model (refer to Section 1 and Section 5.7 for details) and will not be utilized for rumor detection.

PHEME \cite{zubiaga2016analysing} has 2402 conversation threads related to nine events. Following previous studies, we conduct 9-fold cross-validation to obtain robust results. In each fold, conversation threads related to one event are used for testing, and all the conversations related to the other eight events are used for training. 

Both datasets have the problem of class distribution imbalance, so we choose Macro-$F_1$ as the primary evaluation metric and accuracy as the additional evaluation metric. We run each model five times on both datasets and calculate the average results. 
\begin{table}[tb]
\centering
\caption{Statistics of the datasets}
\begin{tabular}{lcccccc}
\hline
Dataset & \#Thread & \#Tweet & Depth & \#True
rumors & \#False rumors & \#Unverified rumors \\
\hline
SemEval & 325       & 5568     & 3.5   & 145            & 74              & 106                  \\
PHEME   & 2402      & 105354   & 2.8   & 1067           & 638             & 697                  \\
\hline
\end{tabular}
\label{t1}
\end{table}

\subsection{Experimental Settings}
The output sizes of two GCN layers are $d_1$=$d_2$=64. The window size of filters is $t$=1, and the filter number is $K$=64. We set the rate of Dropedge and that of Dropout are both 0.5. We train our model with a 0.0005 learning rate and 64 batch size. We apply an L2-regularizer to the weights of our model to prevent overfitting, and the weight penalty is 0.001. Our proposed model is trained for 100 epochs. The optimizer we used is Adam \cite{kingma2014adam}.
The hyperparameters of baseline methods are referenced from original papers. The source code can be available in Github\footnote{\url{https://github.com/wuyang45/GMVCN-master}}.

\subsection{Baselines}
\begin{itemize}
    \item BranchLSTM \cite{kochkina2017turing}: A architecture that models sequential branches in a conversation thread based on Long Short-Term Memory (LSTM).
    \item TD-RvNN \cite{ma2018rumor}: A rumor detection approach models top-down propagation structure using tree-structured recursive neural networks.
    \item Hierarchical GCN-RNN \cite{wei2019modeling}: A hierarchical framework leverages GCN and RNN to model the structure and temporal property of the threads, respectively.
    \item PLAN \cite{khoo2020interpretable}: A transformer-based model that encodes the conversation thread using a randomly initialized Transformer.
    \item Hierarchical Transformer \cite{yu2020coupled}: A extension model of BERT that first models the interactions in each subthread and then encodes global interactions of all posts based on a Transformer layer.
    \item Bi-GCN \cite{bian2020rumor}: A GCN-based model that learns high-level representations from top-down and bottom-up views of conversation threads.
    \item ClaHi-GAT \cite{lin2021rumor}: A rumor detection model based on GAT that represents the conversation thread as an undirected graph.
    \item EBGCN \cite{wei2021towards}: Variants of Bi-GCN that adaptively adjust weights of unreliable relations by a Bayesian method.
\end{itemize}

\subsection{Main Results}
\begin{table}[]
\centering
\caption{Results of rumor detection.}
\setlength{\tabcolsep}{3mm}{
\begin{tabular}{l|cc|cc}
\hline
\multirow{2}{*}{Method}  & \multicolumn{2}{c|}{SemEval}             & \multicolumn{2}{c}{PHEME}               \\ \cline{2-5} 
                         & \multicolumn{1}{c}{Macro-$F_1$} & Accuracy & \multicolumn{1}{c}{Macro-$F_1$} & Accuracy \\ \hline
BranchLSTM               & \multicolumn{1}{c}{0.491}    & 0.500    & \multicolumn{1}{c}{0.259}    & 0.314    \\
TD-RvNN                  & \multicolumn{1}{c}{0.509}    & 0.536    & \multicolumn{1}{c}{0.264}    & 0.341    \\
Hierarchical GCN-RNN     & \multicolumn{1}{c}{0.540}    & 0.536    & \multicolumn{1}{c}{0.317}    & 0.356    \\
PLAN                     & \multicolumn{1}{c}{0.581}    & 0.571    & \multicolumn{1}{c}{0.361}    & 0.438    \\
Hierarchical Transformer & \multicolumn{1}{c}{0.592}    & 0.607    & \multicolumn{1}{c}{0.372}    & 0.441    \\
Bi-GCN                   & \multicolumn{1}{c}{0.607}    & 0.617    & \multicolumn{1}{c}{0.316}    & 0.442    \\
ClaHi-GAT                & \multicolumn{1}{c}{0.539}    & 0.536    & \multicolumn{1}{c}{0.369}    & 0.556    \\ 
EBGCN                    & \multicolumn{1}{c}{0.639}    & 0.643    & \multicolumn{1}{c}{0.375}    & 0.521    \\ \hline
GMVCN (Ours)           & \multicolumn{1}{c}{\textbf{0.721}}    & \textbf{0.721}    & \multicolumn{1}{c}{\textbf{0.441}}    & \textbf{0.647}    \\ \hline
\end{tabular}}
\label{t2}
\end{table}

Table \ref{t2} shows the performance of our proposed model GMVCN and all baselines on SemEval 2017 and PHEME datasets. Our proposed model GMVCN has significantly superior performance among all the baselines over all metrics across two datasets. 

First, we observe that all structure-based methods (RvNN, Hierarchical GCN-RNN, Bi-GCN, ClaHi-GAT, EBGCN, and our GMVCN) perform better than branchLSTM, which indicates the effectiveness of the structural message-passing mechanism. 
BranchLSTM decomposes one conversation thread along branches of the entire tree, and each branch is encoded by LSTM. 
Since LSTM can only process data with sequential structure, branchLSTM fails to learn high-level representations of rumors from structure information, resulting in its worse performance. 
Unlike branchLSTM, PLAN and Hierarchical Transformer are sequential models with better performance, thanks to the powerful performance of Transformer. 
Transformer breaks through the spatial limitation of sequential structure and can be regarded as a fully connected graph. 
Therefore, mining the structural properties of conversation threads is critical for detecting rumors.

Second, due to integrating various information from multiple directed views of conversation threads, GMVCN, Bi-GCN, and EBGCN beat other structure-based models, except for Bi-GCN in PHEME, which performs slightly worse. RvNN, GCN-RNN, and ClaHi-GAT model conversations from a single view, so that they can only learn single patterns of rumors, limiting their performances. ClaHi-GAT outperforms Bi-GCN and EBGCN in the PHEME dataset. The reason might be the attention mechanisms used in ClaHi-GAT, including graph-level attentions, post-level attentions, and event-level attentions. A variety of complex attention mechanisms can capture more helpful information, which is more conducive to the performance of ClaHi-GAT on data of real-world scenarios, i.e., PHEME, where train sets and test sets are neither balanced nor correlated. 

Third, GMVCN is significantly superior to Bi-GCN and EBGCN, which demonstrates that indicative features learned in multiple views fusion help improve the performance of rumor detection. Bi-GCN and EBGCN merely concatenate feature representations of multiple views as the final representation of a conversation without deeply mining inter-view complement. However, different views represent the conversation thread from different angles. Just like in spatial geometry, the shape of an object can only be accurately identified by combining three-dimensional views. Hence, the multi-view fusion allows GMVCN to capture clues of rumors from a global perspective, which help improve the performance of our model much more.

All these observations demonstrate that our proposed model GMVCN can effectively detect rumors by integrating the multi-view structural information of the conversation thread.

\subsection{Ablation Study}
\begin{figure}[hp]
	\centering  
	\subfigure[Macro-$F_1$]
	{
		\begin{minipage}[t]{0.47\linewidth}
			\centering
			\includegraphics[width=\textwidth]{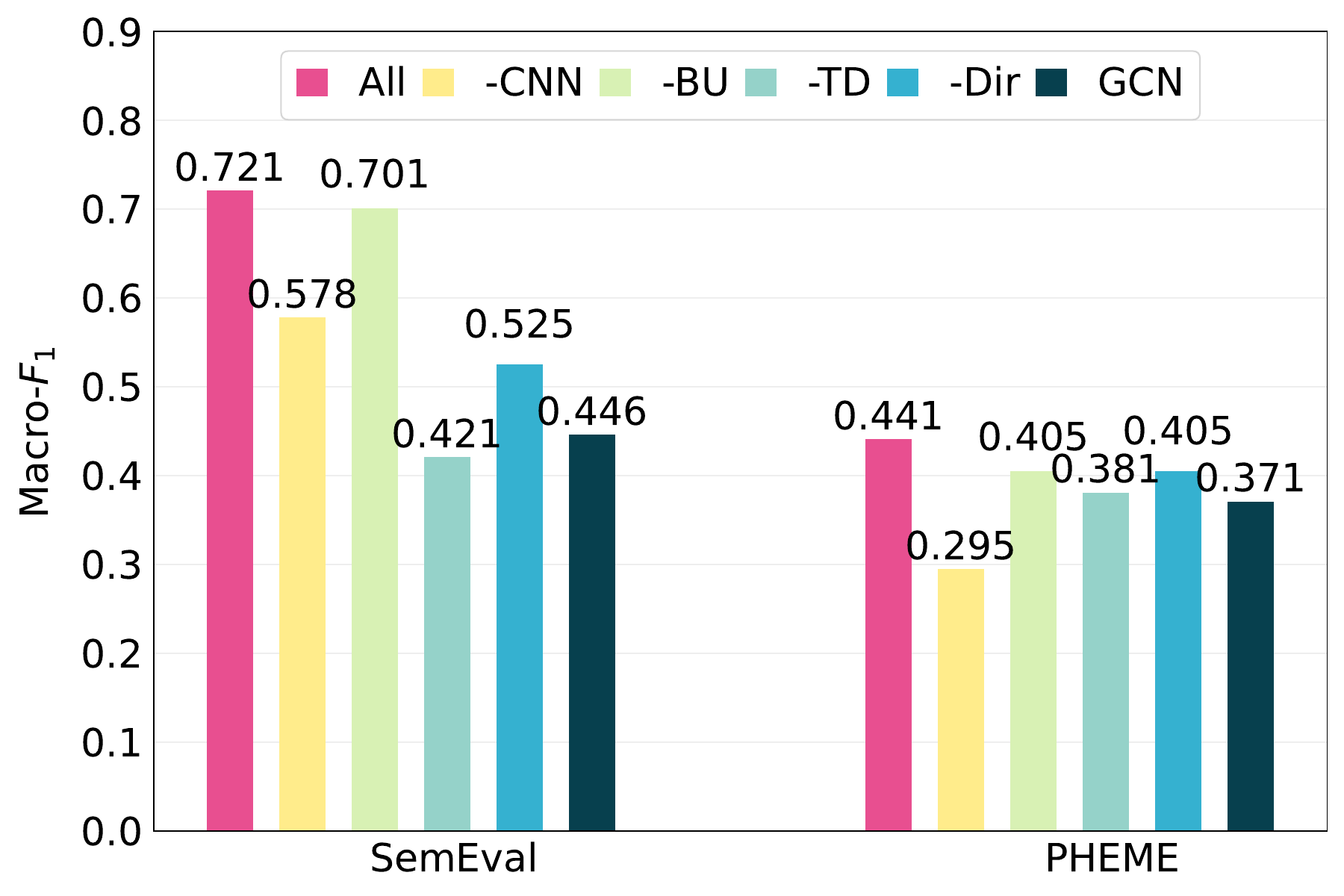}
		\end{minipage}
	}
	\subfigure[Accuracy]
	{
		\begin{minipage}[t]{0.47\linewidth}
			\centering
			\includegraphics[width=\textwidth]{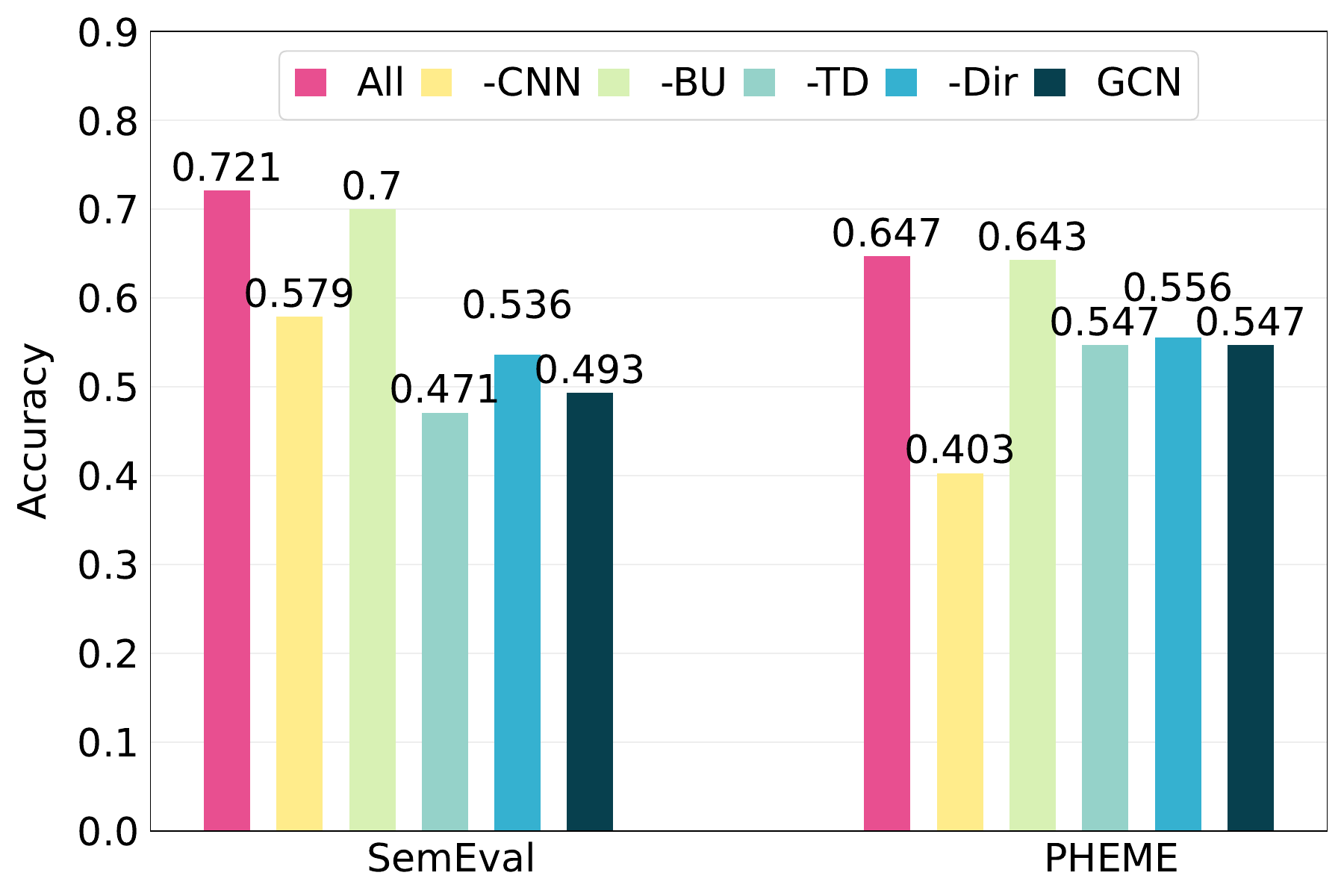}
		\end{minipage}
	} 
	\centering
	\caption{Ablation study of GMVCN on two datasets.}
	\label{f3}
\end{figure}

To analyze the effectiveness of each component of GMVCN, we perform an ablation study in this subsection. “All” denotes the entire mode. We remove each component from the entire model and obtain:
(1)“-CNN”: GMVCN without CNN-based submodule. Like the approaches used in Bi-GCN and EBGCN, we apply the mean-pooling operation on top-down GCN and bottom-up GCN to obtain their representations and then concatenate both features for predicting. 
(2)“-BU”: GMVCN without bottom-up GCN only represents conversations from the top-down view.
(3)“-TD”: GMVCN without top-down GCN.
(4)“-Dir”: We model the conversation thread as an undirected tree structure, which is encoded by a two-layer GCN followed by a CNN submodule.
(5)“GCN”: The vanilla GCN with no direction, i.e., GMVCN without multiple views and inter-view learning.

From Figure \ref{f3}, we draw the following conclusions. First, GMVCN-BU and GMVCN-TD cannot always achieve better results than GMVCN-Dir, but our GMVCN consistently outperforms GMVCN-BU, GMVCN-TD, and GMVCN-Dir in all evaluation metrics. This indicates the superiority of simultaneously considering the top-down and bottom-up views.
Second, the performance of GMVCN-TD drops more significantly than that of GMVCN-BU. It demonstrates that the top-down propagation view can better reflect the characteristics of rumors than the bottom-up dispersion view. Third, after removing the CNN-based component, GMVCN and GMVCN-Dir drop 14.3\%, 7.9\% in Macro-$F_1$ on Semeval, and 14.6\%, 3.4\% in Macro-$F_1$ on PHEME, respectively. There are similar trends in accuracy on both datasets. This proves the effectiveness of CNN in rumor detection, which not only effectively integrates multi-view information but also captures conducive features to identify rumors based on a single view.

\subsection{Parameter Sensitivity}

\subsubsection{Effect of the different window sizes of filters.}
\begin{figure}[htp]
	\centering  
	\subfigure[The window size of filters]
	{
		\begin{minipage}[t]{0.47\linewidth}
			\centering
			\includegraphics[width=\textwidth]{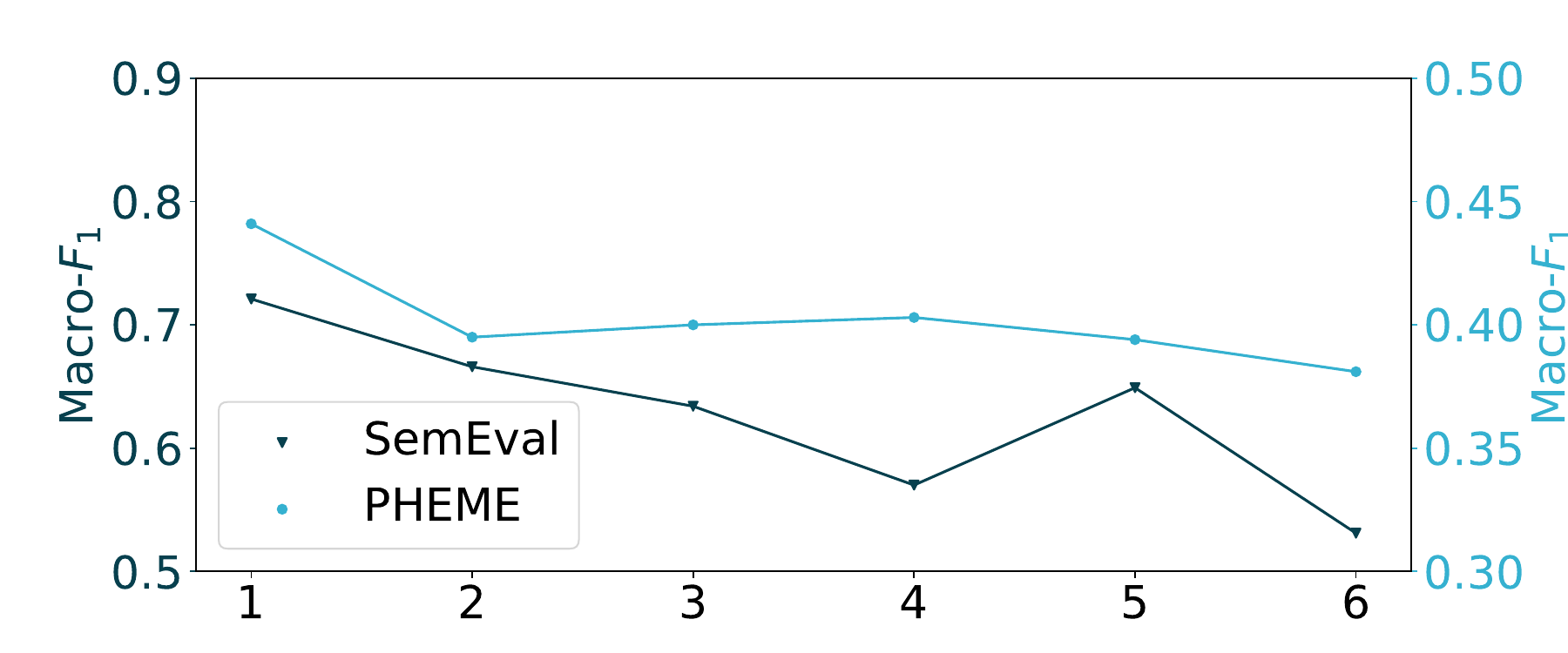}
		\end{minipage}
	}
	\subfigure[The rate of Dropedge]
	{
		\begin{minipage}[t]{0.47\linewidth}
			\centering
			\includegraphics[width=\textwidth]{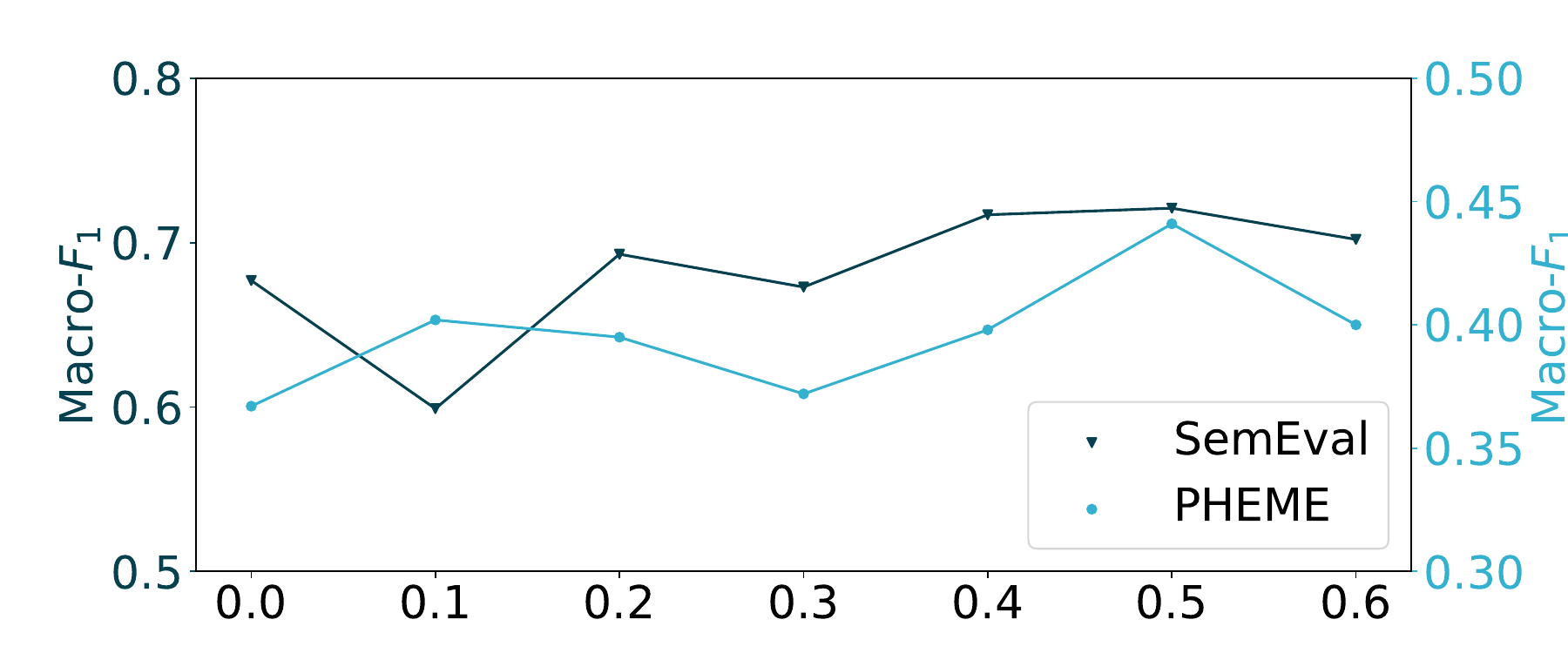}
		\end{minipage}
	} 
	\centering
	\caption{Rumor detection results of GMVCN w.r.t. defferent window size of filters and different rate of Dropedge on two datasets.}
	\label{f4}
\end{figure}

We vary the filter size value in CNN to investigate its influence on rumor detection performance. The results of these experiments are exhibited in Figure \ref{f4}(a). We can observe that our proposed model obtains the best performance on both datasets when the window size of filter is 1. And then, the results are a fall followed by a transitory rise with the increase of the size of filters. This is in line with our intuition. Unlike the correlation between adjacent pixels in an image, there may be no direct correlation between posts in chronological order, e.g., there is no direct connection between adjacent posts (2 and 3, 4 and 5, 5 and 6) of the false rumor conversation thread in Figure \ref{f2}.
Therefore, when the window size is larger than 1, the model learns much noise that affects the performance. However, with the increase in window size, the correlation between users will be slightly enhanced, and the model performance will be improved somewhat. In addition, there are few participating users and data in the early stage of rumor propagation, so the smaller the window, the more conductive the model is to the early detection of rumors. This opens up the possibility of applying our model to early rumor detection.

\subsubsection{Effect of the different rates of Dropedge.}
As shown in Figure \ref{f4}(b), we test the performance of GMVCN by setting hyper-parameter $p$ from 0 to 0.6. The performance rises gradually and reaches the highest point at $p$=0.5. After that, it declines. There exist plenty of unreliable relationships in conversation threads, which leads to severe error accumulation, and weakens model robustness \cite{wei2021towards}. 
With the rate of Dropedge increases, the number of unreliable edges decreases. Although few sound edges are inevitably cut, the increased performance means the model is more robust and can learn more compelling features. However, with further clipping edges, the performance will decline. Our experimental evidence suggests that this reasonable rate can perform the best.

\subsection{Case Study}
\begin{figure}[htp]
	\centering  
	\subfigure[A false rumor]
	{
		\begin{minipage}[t]{0.3\linewidth}
			\centering
			\includegraphics[width=\textwidth]{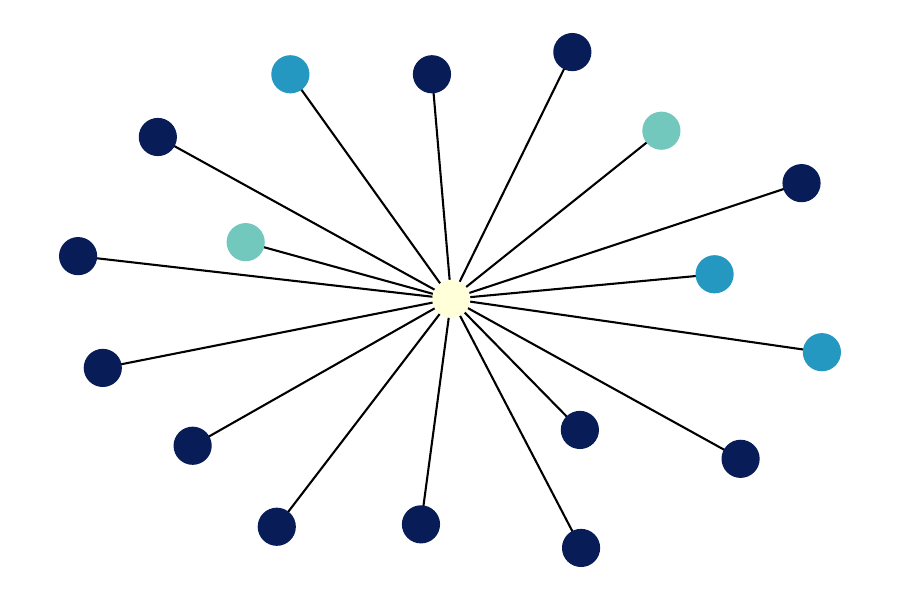}
		\end{minipage}
	}
	\subfigure[A false rumor]
	{
		\begin{minipage}[t]{0.3\linewidth}
			\centering
			\includegraphics[width=\textwidth]{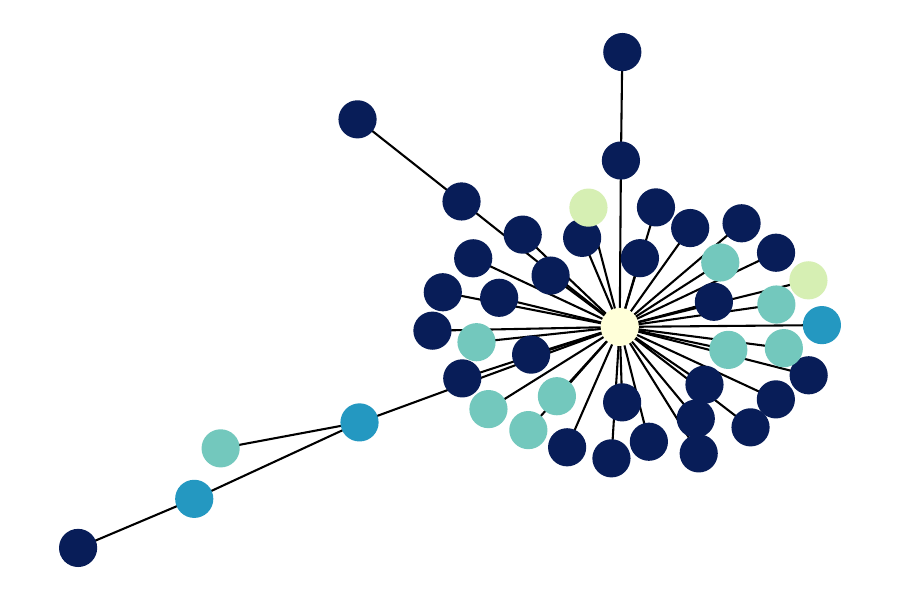}
		\end{minipage}
	} 
	\subfigure[A false rumor]
	{
		\begin{minipage}[t]{0.3\linewidth}
			\centering
			\includegraphics[width=\textwidth]{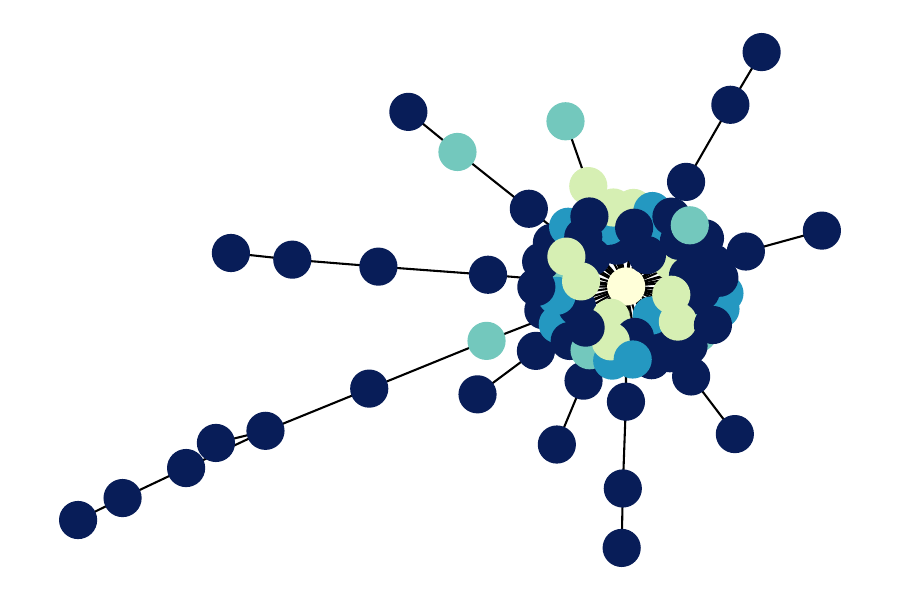}
		\end{minipage}
	}
	
    \subfigure[The feature map of (a)]
	{
		\begin{minipage}[t]{0.3\linewidth}
			\centering
			\includegraphics[width=\textwidth]{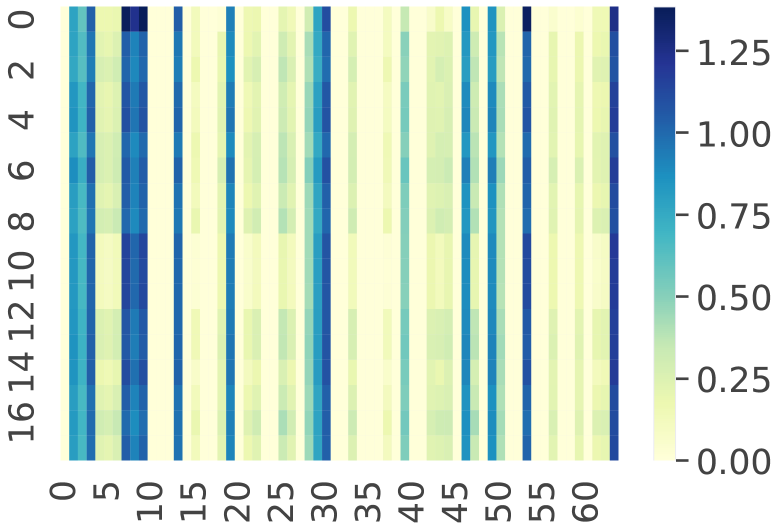}
		\end{minipage}
	}
	\subfigure[The feature map of (b)]
	{
		\begin{minipage}[t]{0.3\linewidth}
			\centering
			\includegraphics[width=\textwidth]{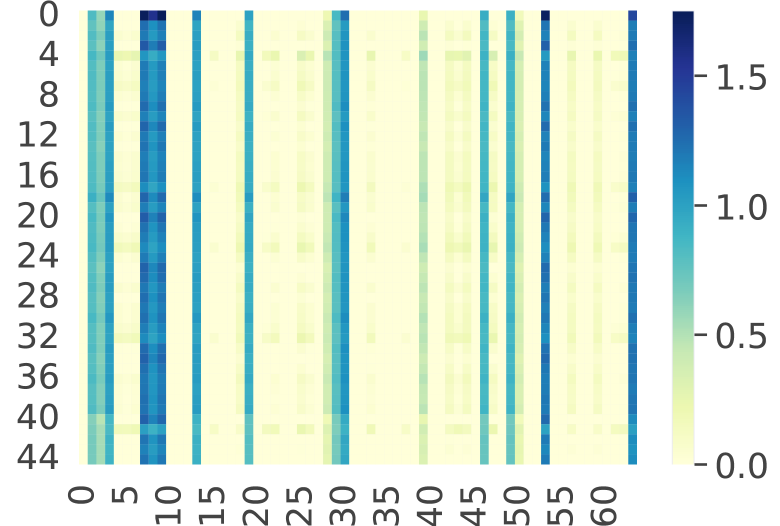}
		\end{minipage}
	} 
	\subfigure[The feature map of (c)]
	{
		\begin{minipage}[t]{0.3\linewidth}
			\centering
			\includegraphics[width=\textwidth]{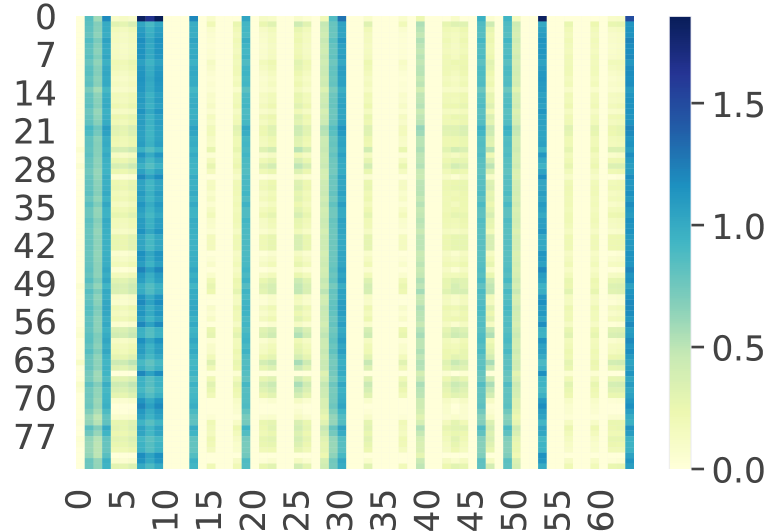}
		\end{minipage}
	}

	\subfigure[A true rumor]
	{
		\begin{minipage}[t]{0.3\linewidth}
			\centering
			\includegraphics[width=\textwidth]{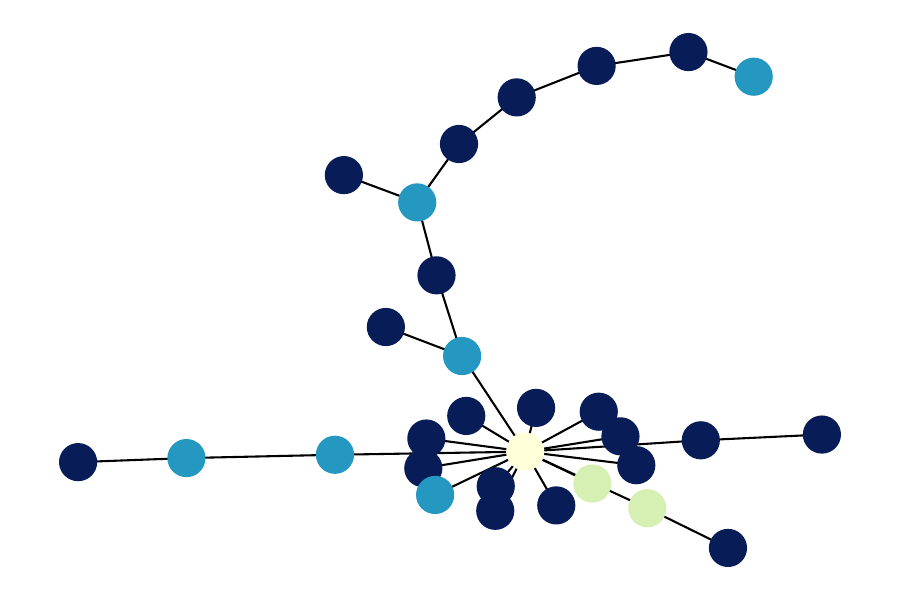}
		\end{minipage}
	}
	\subfigure[A true rumor]
	{
		\begin{minipage}[t]{0.3\linewidth}
			\centering
			\includegraphics[width=\textwidth]{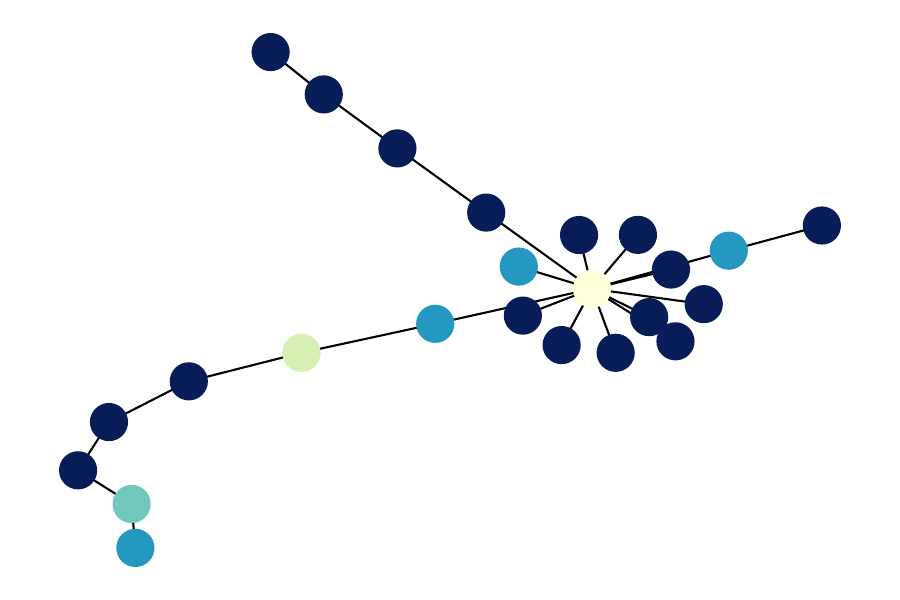}
		\end{minipage}
	}
	\subfigure[A true rumor]
	{
		\begin{minipage}[t]{0.3\linewidth}
			\centering
			\includegraphics[width=\textwidth]{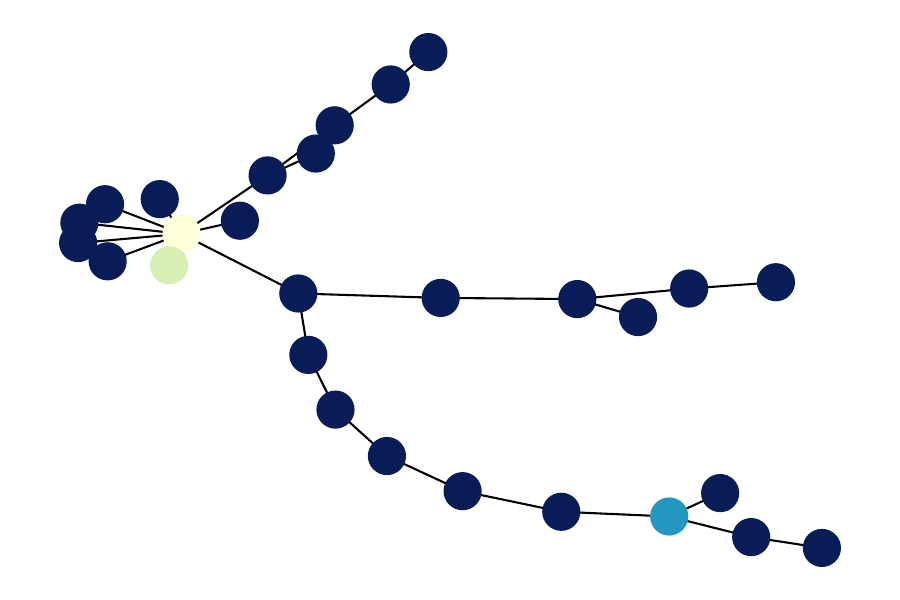}
		\end{minipage}
	}
	
	\subfigure[The feature map of (g)]
	{
		\begin{minipage}[t]{0.3\linewidth}
			\centering
			\includegraphics[width=\textwidth]{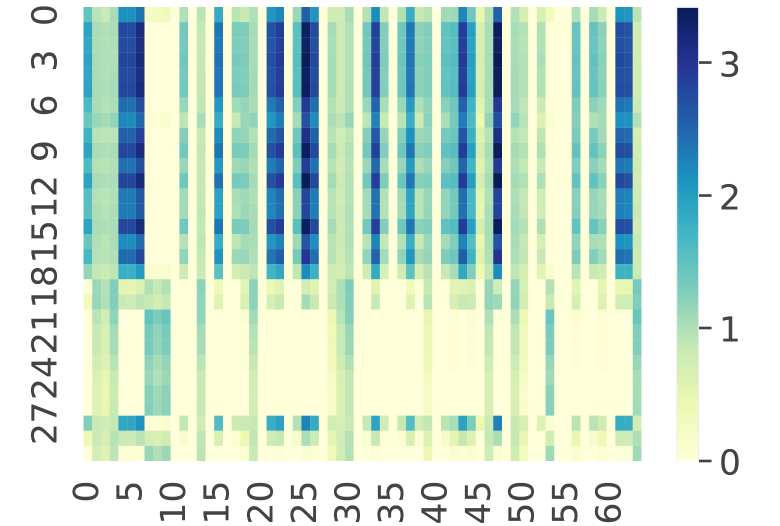}
		\end{minipage}
	}
	\subfigure[The feature map of (h)]
	{
		\begin{minipage}[t]{0.3\linewidth}
			\centering
			\includegraphics[width=\textwidth]{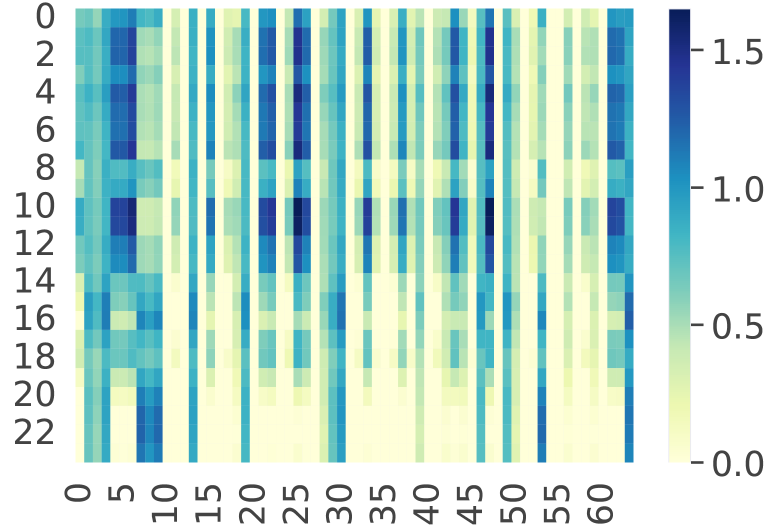}
		\end{minipage}
	}
	\subfigure[The feature map of (i)]
	{
		\begin{minipage}[t]{0.3\linewidth}
			\centering
			\includegraphics[width=\textwidth]{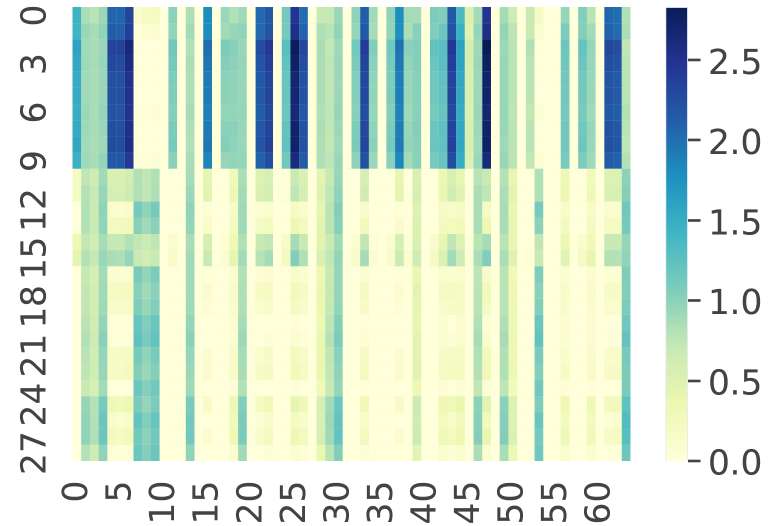}
		\end{minipage}
	}
	\caption{False and true rumors correctly identified by GMVCN and their corresponding feature maps learned by CNN.}
	\label{f5}
\end{figure}

To further illustrate that our proposed model can capture distinguishing features, we visualize feature maps of several representative samples correctly identified by GMVCN from the test set of SemEval to compare the difference between false rumors and true rumors. Feature maps are visualized using heatmap, where Y-axis represents the node index along the time sequence in the propagation conversation (0 is the root node), and the X-axis is the dimension of the feature vector. The larger the feature value, the darker the color. In Figure \ref{f5}, each column is the pair of a conversation graph and its feature map, i.e., [(a)(d)], [(b)(e)], [(c)(f)], [(g)(j)], [(h)(k)], and [(i)(l)]. The meanings represented by the colors of nodes are the same as those in Figure \ref{f1}.

Figure \ref{f5}(a-f) shows three false rumors' conversation topology graphs and their corresponding feature maps learned by the CNN layer of our model. We can see that the three graphs have  simple structures dominated by first-order propagation, and many people disagree with the source claims. Their feature maps are highly similar, where the values of the root node are always larger than those of other nodes. This indicates that the source post of a rumor always plays a vital role in rumor detection, which is consistent with the conclusions of previous work \cite{lin2021rumor,bian2020rumor}.
From figure \ref{f5}(g-l), we can find that true rumors are usually discussed by a small group of people first and then form some long-lasting branch discussions. 
In the three feature maps (j-l), the feature values of early posts are significantly larger than those of later posts, which is utterly different from feature maps of false rumors.

These comparative cases prove that our proposed GMVCN model can notice salience indicators to distinguish false rumors from true rumors by jointly learning textual and structural features. Notably, the feature maps of false (true) rumors are similar, which indicates that our model can learn shared features among all false (true) rumorous events, which is beneficial to improve the performance of rumor detection in newly emerging events. Whether it is a false rumor or a true rumor, there are some common features in their feature maps (i.e., the feature values of all nodes are highlighted in several dimensions). This might be due to the shared features of all event conversations on social media, such as tree topology, words, punctuation marks, etc.

\section{Conclusion}
In this paper, we proposed a novel graph-aware multi-view convolutional neural network to collaboratively integrate the abundance of information from multiple conversation views for rumor detection on social media.
Inspired by computer vision, we regard a rumorous conversation thread as a color image, multi-view graphs as multiple channels, and graph nodes as pixels. 
GCNs in our model could encode the discriminative features of each view.
CNNs in our model could capture the consistent and complementary information over all views and fuse these view-level features to generate the global conversation representation.
Experimental results on two real-world datasets confirm that our GMVCN outperforms state-of-the-art baselines in very large margins.


\section*{Acknowledgments}
This research was supported by National Research and Development Program of China (No.2019YFB1005200).

%
%
%
%


\end{document}